\documentclass[sigconf]{acmart}
\usepackage{multirow}
\AtBeginDocument{%
  }

\setcopyright{acmlicensed}

\copyrightyear{2026}
\acmYear{2026}

\acmConference[CAIN'26]{5th International Conference on AI Engineering – Software Engineering for AI}{April 15--17,2026}{Rio de Janeiro}

\author{Choro Ulan uulu}
\authornote{Corresponding author}
\email{choro.ulan-uulu@siemens.com}
\author{Mikhail Kulyabin}
\author{Iris Fuhrmann}
\author{Jan Joosten}
\author{Nuno Miguel Martins Pacheco}
\author{Filippos Petridis}
\author{Rebecca Johnson}
\affiliation{%
  \institution{Siemens AG}
  \city{Munich}
  \country{Germany}
}

\author{Jan Bosch}
\email{j.bosch1@tue.nl}
\affiliation{%
  \institution{Department of Computer Science and Engineering, Chalmers University of Technology}
  \city{Gothenburg}
  \country{Sweden}
}
\affiliation{%
  \institution{Department of Mathematics and Computer Science, Eindhoven University of Technology}
  \city{Eindhoven}
  \country{Netherlands}
}

\author{Helena Holmström Olsson}
\email{helena.holmstrom.olsson@mau.se}
\affiliation{%
  \institution{Department of Computer Science and Media Technology, Malmö University}
  \city{Malmö}
  \country{Sweden}
}

\title{How to Build AI Agents by Augmenting LLMs with Codified Human Expert Domain Knowledge? A Software Engineering Framework}

\begin{abstract}
Critical domain knowledge typically resides with few experts, creating organizational bottlenecks in scalability and decision-making. Non-experts struggle to create effective visualizations, leading to suboptimal insights and diverting expert time. This paper investigates how to capture and embed human domain knowledge into AI agent systems through an industrial case study. We propose a software engineering framework to capture human domain knowledge for engineering AI agents in simulation data visualization by augmenting a Large Language Model (LLM) with a request classifier, Retrieval-Augmented Generation (RAG) system for code generation, codified expert rules, and visualization design principles unified in an agent demonstrating autonomous, reactive, proactive, and social behavior. Evaluation across five scenarios spanning multiple engineering domains with 12 evaluators demonstrates 206\% improvement in output quality, with our agent achieving expert-level ratings in all cases versus baseline's poor performance, while maintaining superior code quality with lower variance. Our contributions are: an automated agent-based system for visualization generation and a validated framework for systematically capturing human domain knowledge and codifying tacit expert knowledge into AI agents, demonstrating that non-experts can achieve expert-level outcomes in specialized domains.
\end{abstract}

\begin{document}
\settopmatter{printacmref=false}
\maketitle

\section{Introduction}

Organizations across industries face a critical scalability challenge: essential domain knowledge often resides with few experts, creating bottlenecks that limit productivity and decision-making quality \cite{rosen2007overcoming}. When experts are unavailable, work either halts or proceeds with suboptimal outcomes, potentially leading to missed deadlines, increased costs, and catastrophic failures \cite{rosen2007overcoming}.

This challenge is particularly acute in data visualization, where creating effective charts requires both domain knowledge and visualization expertise. Non-experts typically default to familiar chart types because selecting appropriate techniques for complex data remains difficult \cite{grammel2010information}. Even when attempting sophisticated visualizations, results frequently require expert interpretation \cite{Choe2024Enhancing}, while experts must balance mentorship against their primary responsibilities \cite{Vajpayee2024Fostering}.

In simulation data visualization, these challenges intensify. Engineers need dual expertise in simulation analysis and data analytics to create visualizations revealing decision-critical insights. 
Without this expertise, users significantly underutilize available capabilities, missing opportunities to expose key trade-offs \cite{grammel2010information, Choe2024Enhancing}. Critical visualization design knowledge—such as which plot types reveal specific patterns—remains tacit within domain experts, necessitating continuous validation cycles that divert expert resources from high-value tasks.

We illustrate this through Simulation Analysis software, a design space exploration platform that optimizes parameters (e.g., minimizing weight while maximizing strength).
While Simulation Analysis software includes sophisticated post-result analysis capabilities that enable users to visualize complex data sets, this presents an opportunity to enhance user experience through automated guidance. 

While it includes sophisticated visualization capabilities, users require multiple attempts to identify effective visualization types \cite{gadiparthi2024effective}. This trial-and-error approach is time-consuming and discourages full exploration of features that could accelerate design decisions.

Simulation Analysis software serves as an ideal test case for our framework because its extensive visualization capabilities and comprehensive post-processing features are representative of sophisticated engineering software where automated expert guidance can significantly enhance user productivity.

This paper addresses the research question (RQ): How can domain knowledge from human experts be captured, codified, and leveraged to construct Large Language Model (LLM) - based AI agents capable of autonomous expert-level performance?

Our results demonstrate how human expert domain knowledge can be captured to construct LLM-based AI agents to reduce expert bottlenecks. The resulting AI agent enables non-experts to generate expert-level visualizations that match expert-level quality in technical accuracy, visual clarity, and analytical insight, without requiring constant expert involvement.

The contributions of this paper are: 
(1) A systematic software engineering framework for capturing human domain knowledge and engineering an AI agent through complementary strategies: request classifier, RAG for domain-specific code generation, codified expert rules, and visualization design principles, implemented as a reference architecture demonstrating integration of heterogeneous AI techniques with clear separation of concerns.

(2) Empirical evidence from industrial evaluation with 12 evaluators across five scenarios spanning multiple engineering domains (electrochemical, electromagnetic, mechanical systems) demonstrating 206\% improvement in output quality (mean: 2.60 vs. 0.85 on 0-3 scale), with our system achieving expert-level ratings (Mode=3) consistently versus baseline's poor performance (Mode=0 in 4/5 scenarios), and superior code quality with lower variance (SD: 0.29-0.58 vs. 0.39-1.11).

(3) Demonstration that the framework addresses organizational expert bottlenecks by enabling non-experts to generate expert-level visualizations through simple prompts, effectively democratizing domain knowledge and allowing domain experts to focus on specialized tasks requiring unique expertise.

Section 2 provides background and reviews related work, Section 3 describes our research methodology, Section 4 presents expert interview findings, Section 5 presents our solution, Section 6 validates effectiveness, and Section 7 concludes, and Section 8 discusses limitations.

\section{Background and Related work}

Poor visualizations cause fundamental misinterpretation of critical data \cite{Franconeri2021The}. Medical-risk visualizations can lead patients to fundamentally misunderstand the base rates or risk factors for diseases or medical procedures \cite{ancker2006design, Franconeri2021The}. Suboptimal visualizations in minimally invasive surgery have contributed to patient injury rates exceeding 50\% \cite{ameerah2025blurred}. Misleading visualizations can lead to spread of misinformation \cite{biselli2025chartchecker, lisnic2023misleading}. In the context of simulation software, according to \cite{Kido2023Visualization}, misused visualization types are difficult to interpret and may lead to erroneous decision making.

The challenges in creating effective visualizations are established. Non-experts struggle with selecting appropriate visualization techniques for complex data \cite{grammel2010information}, often defaulting to familiar but suboptimal chart types. Even when more sophisticated visualizations are attempted, they frequently require expert interpretation \cite{Choe2024Enhancing}, creating dependencies on scarce expert resources \cite{Vajpayee2024Fostering}.

\textbf{Knowledge Codification and Rule-Based Systems}
The integration of structured knowledge into LLM systems represents an active research area. \cite{yang2025distilling} conducted preliminary research on how LLMs can learn from rules. \cite{wang2024can} developed a rule generation framework, creating 8000 primitive rules and 6000 compositional rules across five domains. \cite{vertsel2024hybrid} applied rule-based results to an LLM to generate insights. However, earlier work suggested limitations in LLMs' ability to follow rules \cite{mu2023can}, though newer LLM versions have demonstrated significantly improved rule-following capabilities. More recently, \cite{zhu2023large} showed that LLMs with rules perform 30\% better than LLMs without rules, providing empirical evidence for the value of rule integration.

\textbf{LLM based scientific visualization}
We define an AI agent following \cite{Deng2024AI} as a system exhibiting four key properties: autonomy (operates independently after prompting), reactivity (responds to user requests), proactivity (applies expert rules), and social ability (natural language interaction). These characteristics manifest through classifier-based routing, autonomous rule application, and code generation.

Recent advances in Large Language Models have opened new possibilities for automated code generation and visualization creation. \cite{vazquez2024llms} evaluated whether LLMs are ready for generating code that creates visualizations, showing that a high number of charts were properly built and that LLMs are a promising approach to generate charts. \cite{das2025charts} demonstrated that modern multimodal LLMs can surpass human performance in visualization literacy tasks when given the proper analytical framework. Several approaches have emerged for LLM-based visualization generation. \cite{yang2024matplotagent} created a method for LLM Agents-based visualization of data, generating code for the Python library matplotlib and incorporating a feedback loop to validate results. \cite{goswami2025plotgen} employed a multi-agent system to generate plots. \cite{mallick2024chatvis} used example code snippets specific for ParaView's PvPython API to help the LLM generate scripts for 2D visualization creation. The application of AI in engineering software contexts continues to evolve \cite{ulan2026ai}.

\section{Research Method and Case Context}
\textbf{Case Company Context}
This research was conducted at Siemens, focusing on their Simulation Analysis software. The first author's position as a Siemens employee provided privileged access to domain experts, proprietary data, and technical documentation, enabling deep understanding of real-world visualization challenges and facilitating the iterative framework development process.

\subsection{Overall Research Process}
Case study methodology was employed to investigate knowledge codification through a systematic four-step research process. The research followed a sequential approach designed to extract, codify, and validate expert knowledge integration into LLMs.

\subsubsection{Step 1: Expert Knowledge Extraction}
We conducted semi-structured interviews with two domain experts to extract specialized knowledge: a simulation analysis software expert and a data visualization expert. The interviews aimed to understand current challenges, identify knowledge gaps, and extract actionable rules for improving visualization creation in Simulation Analysis software. 

\textbf{Interview Design and Conduct} The interview protocol was structured around three main themes: (1) current visualization workflows and pain points, (2) expert decision-making processes when creating visualizations, and (3) specific rules and heuristics used in practice. Each interview lasted approximately 60-90 minutes and was conducted individually to avoid groupthink effects. The interviews were screen-recorded with participant consent and subsequently transcribed for analysis.

The interview guide included open-ended questions such as "Can you walk me through your typical process when creating a visualization for simulation data?" and "What are the most used and useful plots?" followed by probing questions to elicit specific rules and decision criteria. We used a combination of direct questioning and scenario-based discussions where experts were presented with example visualization tasks to reveal their tacit knowledge.

\textbf{Rule Extraction and Implementation} The experts directly provided explicit rules and guidelines during the interviews, which were captured and documented. Rather than requiring interpretive analysis, the experts articulated clear, actionable rules that could be directly implemented. These expert-provided rules were then systematically codified into Python functions and integrated into our system architecture (Section \ref{solution}). This direct rule provision approach ensured high fidelity between expert knowledge and system implementation, forming the foundation for the framework development in Step 2.

Using two experts is methodologically appropriate for this study because: (i) we require comprehensive coverage of two distinct, specialized knowledge domains that rarely overlap in single individuals, (ii) our goal is systematic knowledge extraction rather than statistical generalization, and (iii) validation relies on objective technical performance metrics rather than expert consensus. Additionally, the knowledge extraction process was significantly enhanced by the first author's position as an employee within the case company, providing excellent access to internal data, company internal discussions, and a variety of proprietary data sources that complemented the formal expert interviews. This approach aligns with established knowledge engineering practices where small expert groups (1-3 experts) are sufficient to construct a coherent knowledge base \cite{hoffman1995eliciting}.

\subsubsection{Step 2: Framework Development}
Following knowledge extraction, we analyzed the interview findings to derive a systematic framework for codifying expert knowledge into AI systems. This step involved identifying common patterns across expert insights, structuring the knowledge into implementable components, and designing a generalized framework that could be applied beyond the specific case study domain.

\subsubsection{Step 3: System Implementation}
We developed an automated system that generates Python code for visualization creation, incorporating the extracted expert knowledge through multiple augmentation mechanisms. The implementation integrated a RAG system for domain-specific code generation, codified expert rules as executable components, enhanced system prompts with visualization guidelines, and unified these elements through an intelligent classifier within a comprehensive architecture.

\subsubsection{Step 4: Comprehensive Validation}
To validate our system, we designed a comprehensive evaluation methodology that assessed both the quality of generated visualizations and the underlying code quality across engineering domains. The validation process involved two key evaluation components:

\textbf{Quality of Final Visualization Assessment} We recruited a non-expert user from within the case company—a mechanical engineer with 1 year of experience in simulation domain but limited visualization expertise—to test the system's practical effectiveness. The participant was introduced to the system through a 30-minute orientation session covering basic functionality without visualization training. The non-expert user generated visualizations using simple prompts, which were then assessed by a visualization expert—a data visualization specialist with 20 years of experience in simulation data analysis, currently working at the case company. The expert evaluated both analytical insight and visual effectiveness based on their professional expertise. This evaluation component assessed whether our system could enable non-experts to generate expert-level visualizations.

\textbf{Quality of Generated Code Assessment} We conducted a technical evaluation of the underlying Python code quality, comparing our proposed system (LLM based AI agent) against a baseline approach (LLM with RAG only) across multiple representative scenarios. We employed a triangulated assessment approach involving two domain experts—a software engineering expert and a simulation optimization expert—supplemented by an AI assessor (Claude 4.5 Sonnet) to enhance evaluation robustness and reduce potential bias. The evaluation focused on three critical dimensions: code validity, code correctness, and output quality.

\section{Expert Knowledge Extraction and Analysis}

We conducted semi-structured interviews with two distinct domain experts to extract specialized knowledge from different areas of expertise. The first interview focused on simulation analysis domain knowledge, while the second concentrated on visualization design principles.

\subsection{Simulation Analysis Software Expert Insights}
\label{simins}

Our interview with the simulation analysis software domain expert revealed four primary challenges in simulation data visualization: (1) Expert bottleneck: Significant expert time is required to guide users through effective visualization creation, creating scalability constraints. (2) Advanced feature adoption: Maximizing the value of sophisticated visualization capabilities requires domain knowledge that can be systematically captured and transferred to users. (3) Iterative complexity: Creating decision-enabling visualizations requires extensive refinement to identify critical insights and effective visual encodings.

\subsubsection{Simulation Analysis software}
\label{simsofrules}
\textbf{Domain expert rules for plot generation}
During the interview, a set of rules was obtained that guide users through the post-processing workflow in Simulation Analysis software. An example of a rule is convergence of objectives. The study is not finished if the objective is not converged.

A key insight from the expert was that Simulation Analysis software operates as a physics-agnostic platform, meaning it can be applied across diverse engineering domains without being tied to specific physical phenomena or simulation types.

The expert emphasized that the rules should mirror this physics-agnostic nature of Simulation Analysis software. Consequently, these guidelines were deliberately formulated to be domain-independent, because the domain expert works with users that are for example working with: (i) Computational Fluid Dynamics (CFD) simulations. (ii) Pressure drop analyses. (iii) Stress analysis studies.

This universal applicability is crucial because it allows the same post-processing methodology to be consistently applied across different engineering disciplines, reducing the learning curve for users who work in multiple domains.

\textbf{Example rule: Convergence Assessment}
Any analysis begins with a fundamental question: Have the objectives converged? This initial step is critical because: it validates the reliability of the optimization results, it determines whether additional iterations are needed and it provides confidence in subsequent analysis steps.

To answer this question, users are guided to create a history plot that visualizes the objective function values over the course of the optimization. This graphical representation allows users to clearly observe whether the objectives have reached a stable, converged state or if the optimization is still actively searching for better solutions.

\subsection{Visualization Expert Insights}
\label{vizins}
In a separate interview, we consulted with a visualization design expert to extract principles for creating effective and readable visualizations. This expert provided complementary knowledge focused on visual design and communication effectiveness.

\subsubsection{History Plot Requirements}

The visualization expert provided specific guidelines for history plots in simulation contexts. History plots should always display the best design for reference and limit displays to no more than 2 variables, objectives, or responses to maintain readability. Visual encoding should communicate convergence status through dashed lines for non-converged variables and solid lines for converged variables. Related objectives should be grouped on the same plot for comparison, establishing clear visual hierarchy with primary and secondary objectives. Both objectives must show their individual best design history while prioritizing readability over decorative styling.

\subsubsection{General Visualization Principles:}
The expert emphasized that effective simulation visualizations must: clearly communicate convergence status through visual encoding, maintain visual clarity and avoid unnecessary styling that reduces readability, provide sufficient context for interpretation without expert guidance and follow consistent styling patterns across different plot types.

These insights revealed that technical correctness alone was insufficient - visualizations needed to follow established design principles to effectively communicate insights to non-expert users. Additionally, according to \cite{lin2021fooled} beautiful visualizations indicate increased trust.

\subsection{Summary of Key Findings}
Based on the combined insights from both expert interviews and the first authors position as an employee within the case company, we derived a framework described in Section \ref{framework}, which is validated in this paper by an implementation of the solution to the challenges.

\subsection{Software Engineering Framework}
\label{framework}

\subsubsection{Framework Purpose and Derivation}

The expert bottleneck in data visualization represents a broader organizational challenge where critical domain knowledge remains concentrated among few specialists. Our framework systematically codifies tacit expert knowledge into AI systems, enabling automated replication of expert-level decision-making. The framework addresses three core challenges: (1) capturing diverse forms of expert knowledge (explicit rules vs. tacit guidelines), (2) implementing this knowledge in forms that AI systems can effectively utilize, and (3) ensuring the resulting systems maintain expert-level quality without requiring expert oversight.

This framework emerged directly from analyzing expert interview transcripts (Section \ref{simins}). A key pattern emerged: experts described two fundamentally different types of knowledge:

\textbf{Explicit procedural rules} As described in Section \ref{simins}, the simulation expert emphasized convergence checking as a prerequisite for analysis. This type of knowledge—clear if-then logic—could be directly translated into executable code (Python functions that check convergence and generate reports).

\textbf{Tacit design principles} The visualization expert's guidelines on visual encoding (e.g., using dashed lines for non-converged variables, Section \ref{simins}) represent contextual, judgment-based knowledge that couldn't be reduced to simple rules. This required integration into the AI's reasoning process through prompt engineering or retrieval mechanisms.

This insight led to our two-pronged implementation strategy: (1) executable code for algorithmic rules, and (2) LLM augmentation for contextual principles. Initial testing revealed neither approach alone was sufficient—code lacked analytical insight while LLM-only solutions produced domain-inappropriate outputs. This drove the integrated approach where both strategies work complementarily.

\subsubsection{Framework Overview}
\begin{figure}
    \centering
    \includegraphics[width=1\linewidth]{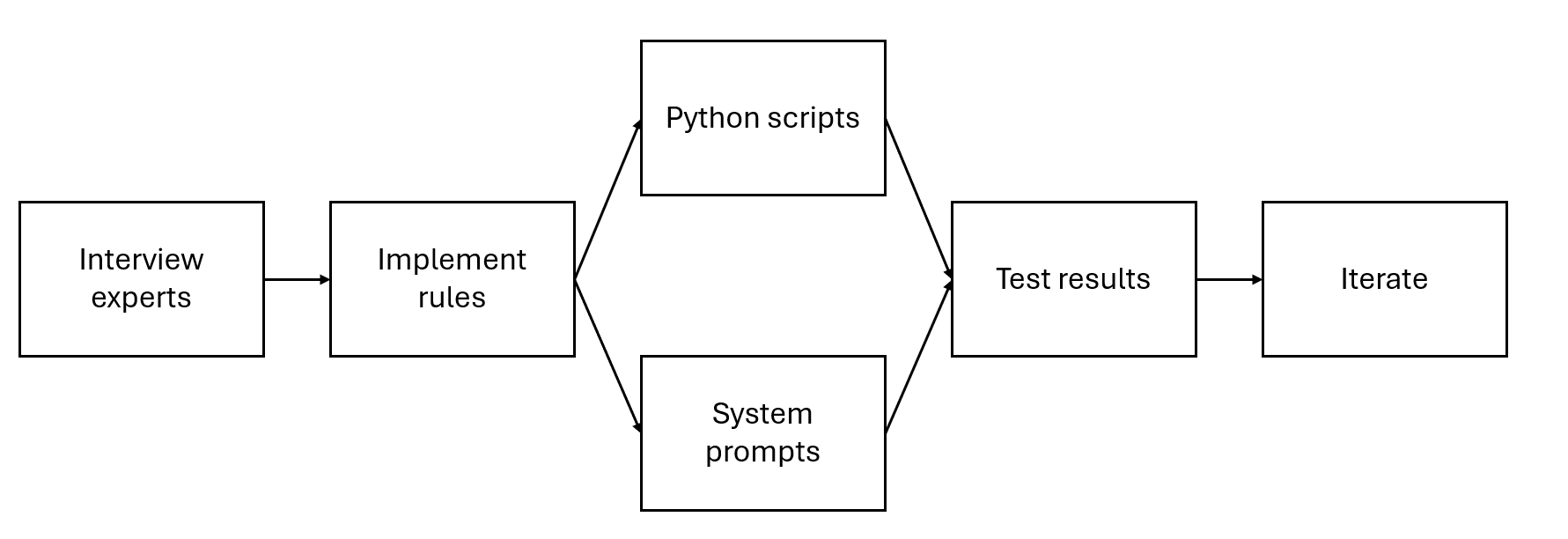}
    \caption{Software Engineering Framework for Knowledge Codification}
    \label{fig:framework}
\end{figure}

We developed a four-step framework (Fig. \ref{fig:framework}) to codify expert knowledge: (1) conduct expert interviews, (2) implement rules through complementary approaches—writing Python scripts for algorithmic rules and embedding guidelines directly into LLM system prompts, (3) test results, and (4) iterate based on findings.

This validated framework (Section \ref{solution}) provides a systematic approach for transferring expert knowledge into an AI agent.

\section{Solution}
\label{solution}

\subsection{Framework Validation through Implementation}

\begin{figure}
    \centering
    \includegraphics[width=1\linewidth]{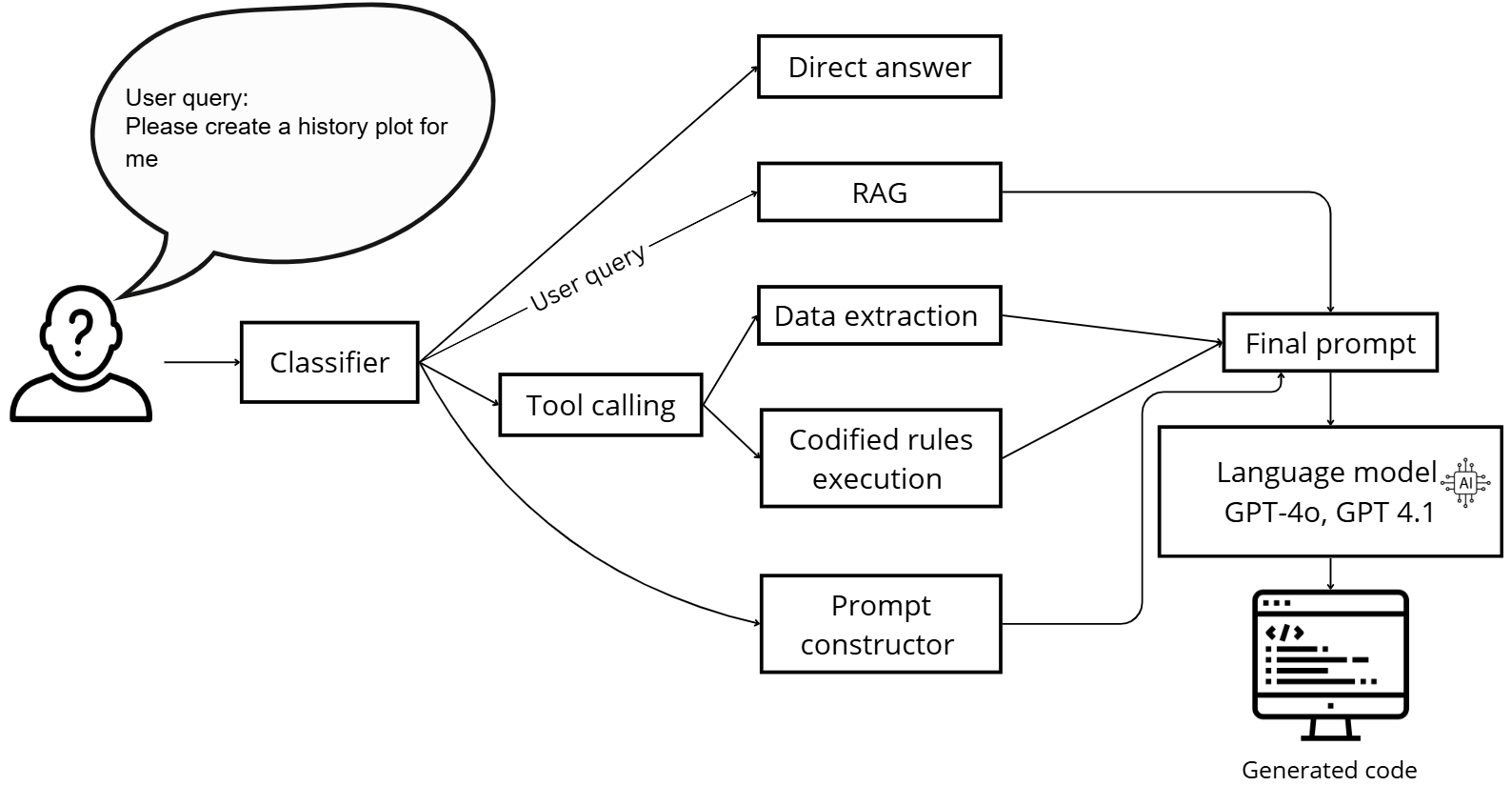}
    \caption{AI agent architecture. User query enters the agent's perception layer (classifier), which autonomously routes to appropriate processing components.}
    \label{fig:pipeline}
\end{figure}

To validate our framework, we developed a comprehensive AI Agent (Fig. \ref{fig:pipeline}) that automates expert-level visualization generation by integrating a classifier using codified expert rules, prompt constructor (for visualization guidelines and code generation) from domain experts, and a RAG system.

The agent architecture begins with intelligent request classification, where a classifier-based system categorizes user requests and dynamically invokes appropriate processing scripts based on predefined expert rules, ensuring efficient routing of diverse query types. At the core, an LLM connected to a RAG system containing domain-specific knowledge—including code examples and technical manuals—generates Python code that produces targeted visualizations when executed within the simulation analysis environment.

Domain expert knowledge is codified into executable rules that generate analytical reports, which are automatically integrated into the system prompt to enhance the LLM's understanding of domain-specific requirements and industry best practices. When initial results revealed that generated plots lacked informativeness and contained visual flaws (Fig. \ref{fig:2_relation_bad}), we extracted additional rules from visualization experts focusing on the most frequently used plot types: 2D relation plots, parallel plots, and history plots (Section \ref{vizins}). These expert-derived rules were directly implemented in the system prompt to guide visualization generation.

The final component combines all previous outputs into a comprehensive query incorporating user requirements, domain knowledge, and visualization best practices. Based on this enhanced query, the LLM generates optimized scripts that produce high-quality visualizations. This implementation addresses critical challenges through intelligent classification for real-time responsiveness, codified rules and RAG for domain expertise integration, expert-derived guidelines for visual quality assurance, and scalable architecture supporting diverse visualization requirements.

\section{Technical validation}
\label{sec:validation}

To validate our agent-based approach, we conducted a comparative analysis to demonstrate its effectiveness in mitigating the expert bottleneck. We assess our agent from two perspectives:
\textbf{1. Quality of the final visualization} 
1. We show that our AI agent enables a non-expert to generate visualizations that are as insightful as those created by a human expert, using only a simple prompt through autonomous agent operation. 
\textbf{2. Quality of the Generated Code} 
2. We provide a technical evaluation of the underlying Python code, proving that our system generates objectively better code by correctly implementing codified expert knowledge.

Our validation compares a \textbf{Baseline System} (a LLM only connected to RAG) against our \textbf{Proposed System} (LLM based AI agent). We demonstrate that our agent produces expert-level results without requiring expert involvement from the user.

\subsection{Quality of the Final Visualization}

We present scenarios where non-expert users generate insightful plots from their Simulation Analysis software simulation data. 
A visualization expert—a data visualization specialist with 20 years of experience in simulation data analysis, currently working at the case company — evaluated the generated plots.
The expert was provided with the original data, the user's request, and the generated visualization, and asked to evaluate both analytical insight and visual effectiveness based on their professional expertise and experience with simulation data visualization.

\subsubsection{Scenario: Visualizing Convergence in a History Plot}

\textbf{Goal} An engineer wants to demonstrate that simulation objectives have converged, a critical step for validating results.

\textbf{User Input (provided to both systems)} \texttt{"Please generate a history plot to check convergence."}

\begin{figure}
    \centering
    \includegraphics[width=1\linewidth]{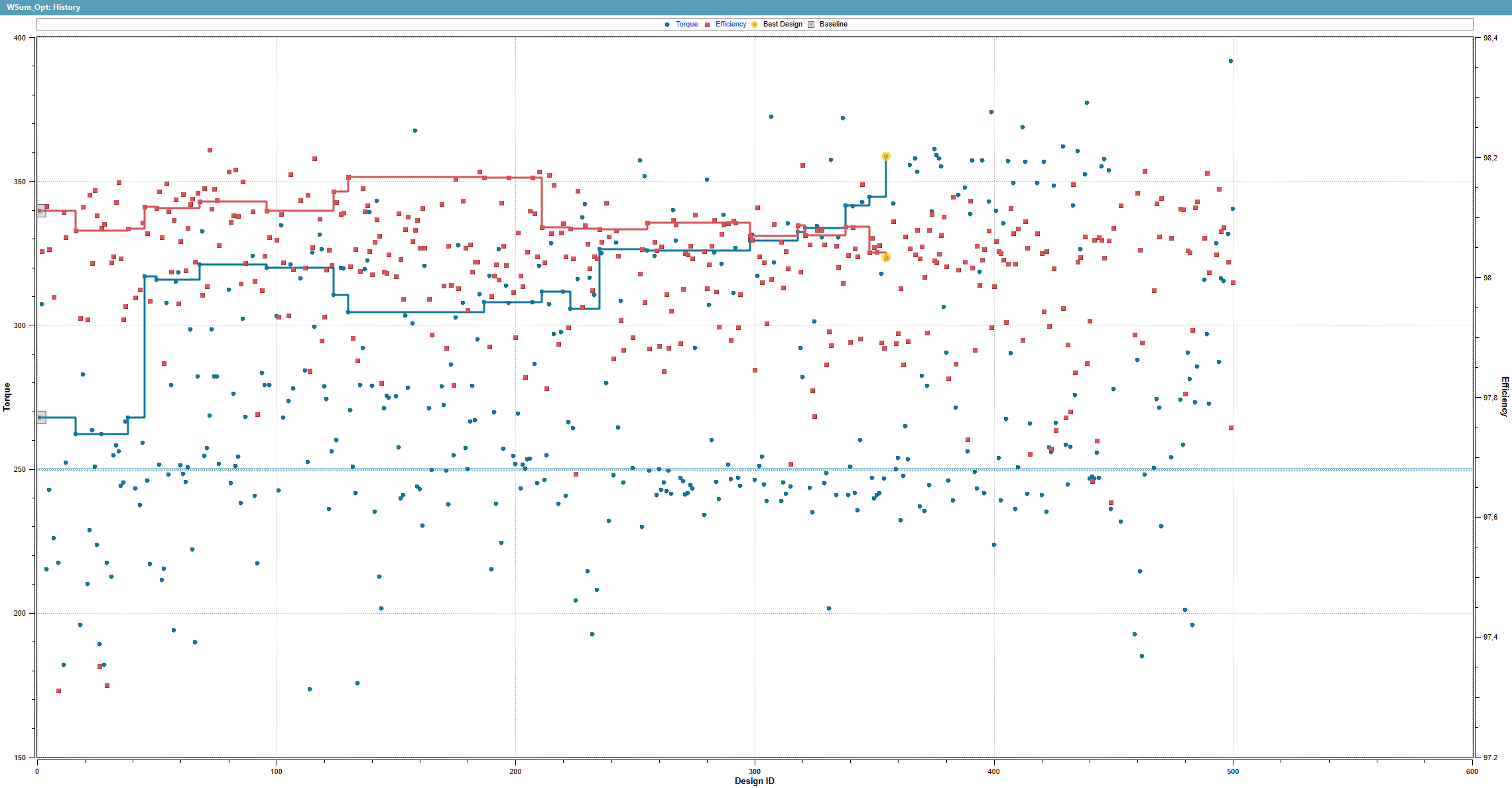}
    \caption{Human generated history plot}
    \label{fig:baseline_history}
\end{figure}

\textbf{Human Result:} The engineer used identical solid lines for both converged and non-converged objectives (Fig. \ref{fig:baseline_history}), failing to communicate convergence status—the primary goal.

\begin{figure}
    \centering
    \includegraphics[width=1\linewidth]{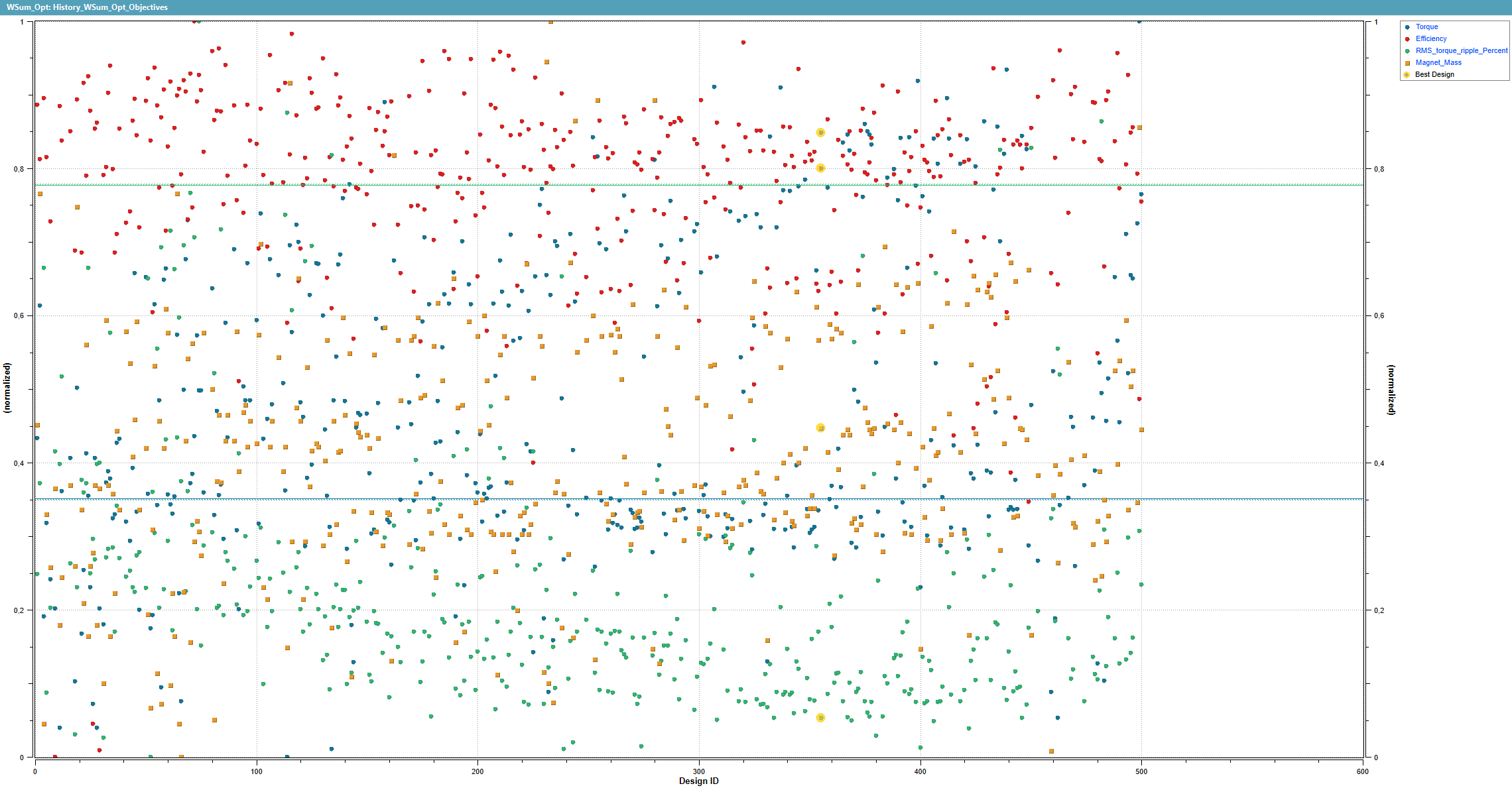}
    \caption{History plot generated by LLM+RAG}
    \label{fig:placeholder}
\end{figure}

\textbf{Baseline System Result:} The LLM+RAG generated a plot without useful information (Fig. \ref{fig:placeholder}).

\begin{figure}
    \centering
    \includegraphics[width=1\linewidth]{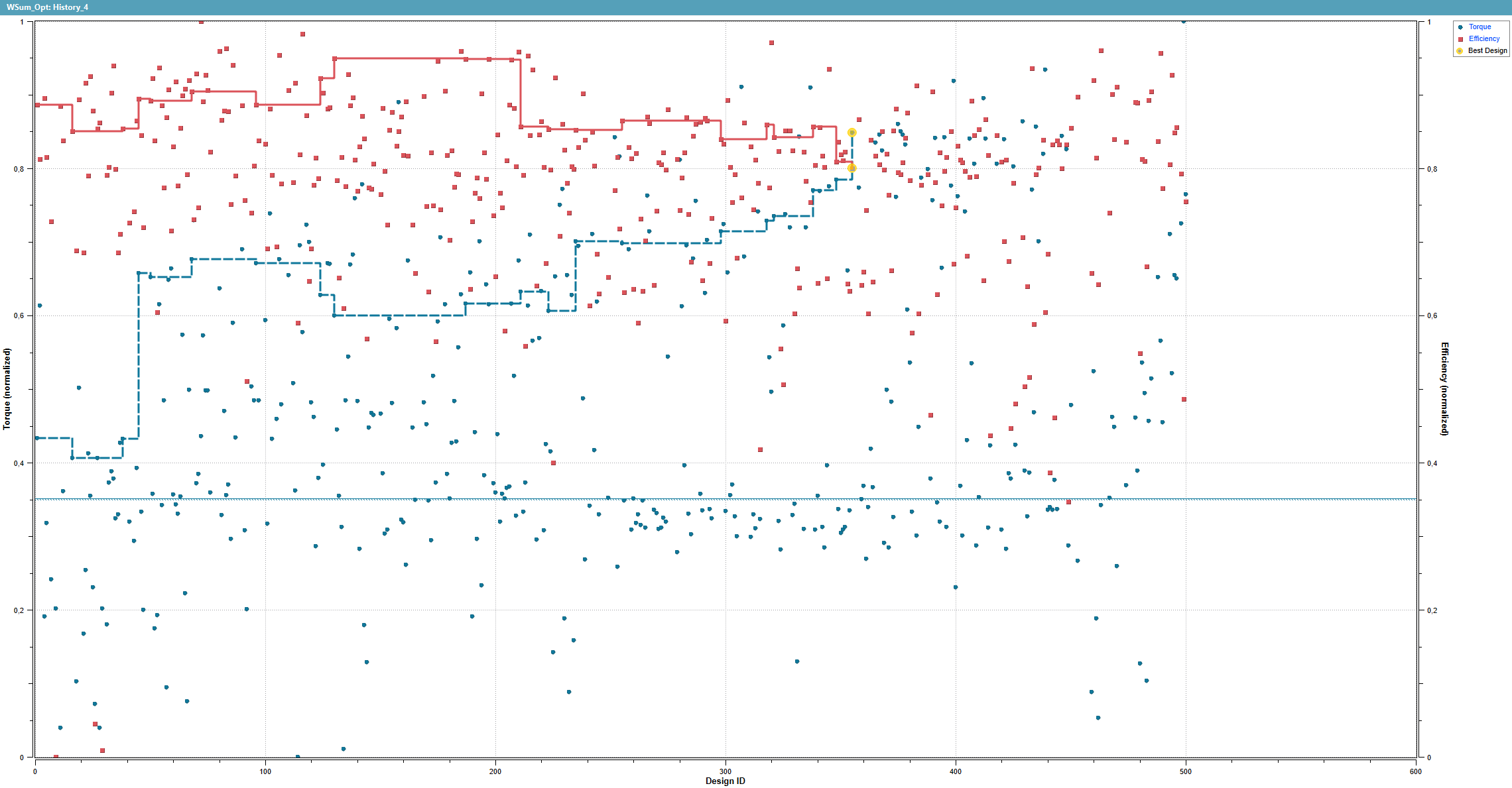}
    \caption{History plot generated by AI agent}
    \label{fig:augmented_history}
\end{figure}

\textbf{Proposed System Result:} Agent correctly applied expert rules, using dashed lines for non-converged and solid lines for converged variables (Fig. \ref{fig:augmented_history}), immediately communicating the status and fulfilling the request accurately.

\subsubsection{Scenario: Creating an Informative 2D Relation Plot}

\textbf{Goal} A junior engineer needs to understand the trade-off between "total mass" and "first torsion frequency" to identify optimal design choices.

\textbf{User Input (provided to both systems)} \texttt{please generate python code for 2d relation plot with variables total mass, first torsional frequency and total cost}

\begin{figure}
    \centering
    \includegraphics[width=1\linewidth]{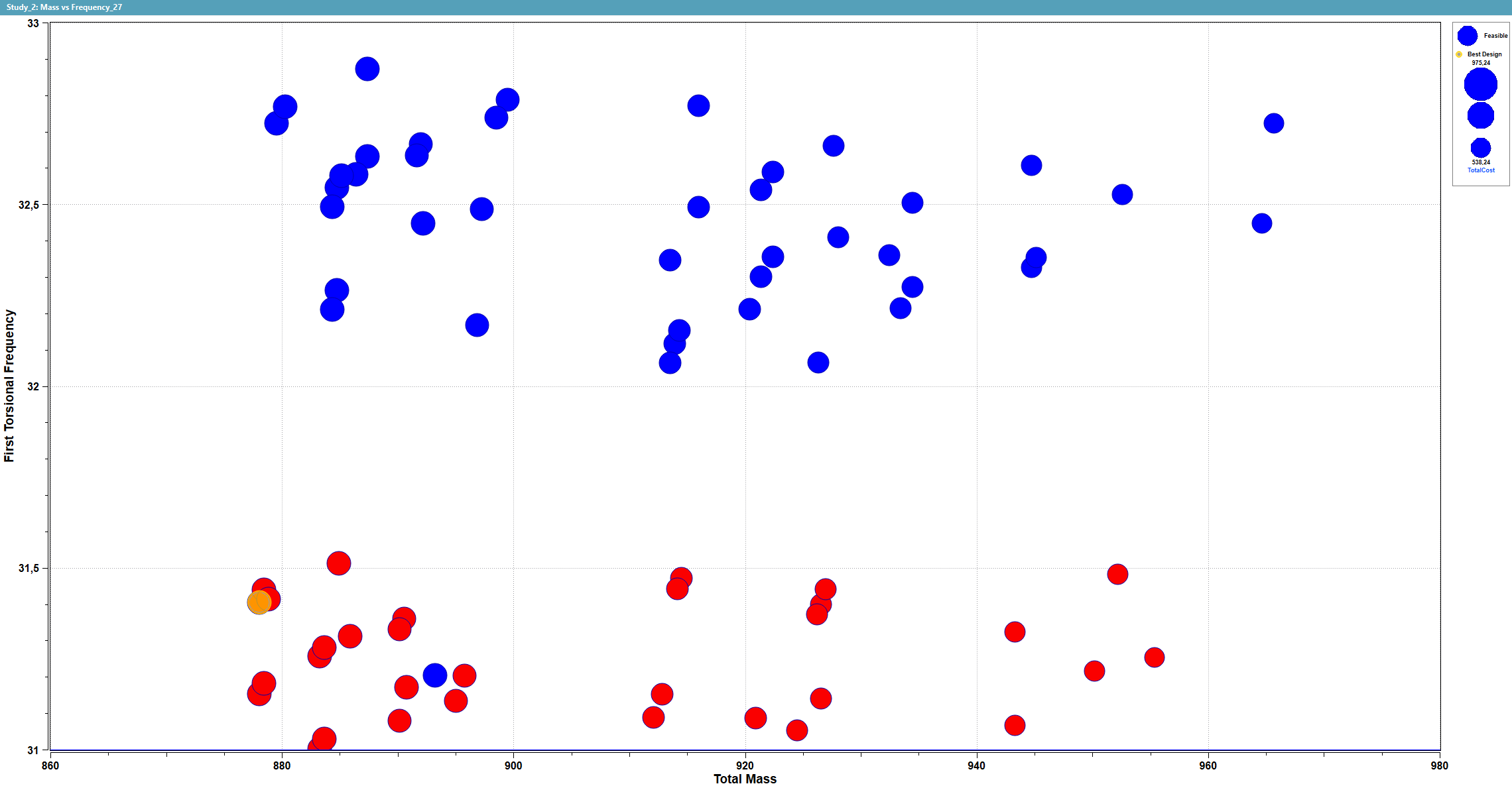}
    \caption{Human generated 2D relation plot}
    \label{fig:2_relation_human}
\end{figure}

\textbf{Human Result:} The expert used color to represent material type and structured the plot to highlight the trade-off between weight and cost (Fig. \ref{fig:2_relation_human}). The plot guides users to identify a cost-effective steel option (blue dot) among lighter aluminum alternatives (red dots).

\begin{figure}
    \centering
    \includegraphics[width=1\linewidth]{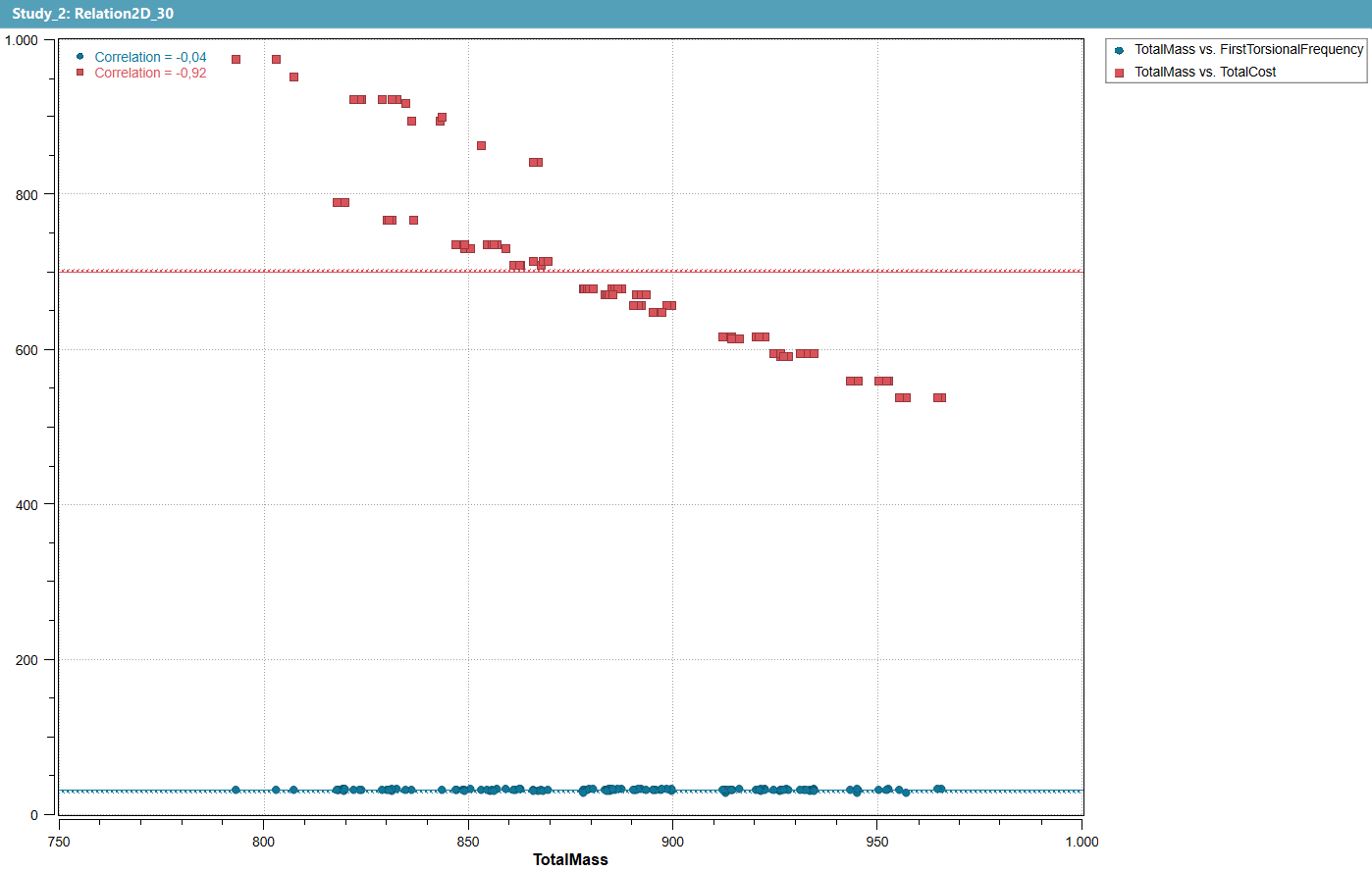}
    \caption{LLM+RAG generated 2D relation plot}
    \label{fig:2_relation_bad}
\end{figure}

\textbf{Baseline System Result:} The baseline LLM produced a technically correct but uninformative plot (Fig. \ref{fig:2_relation_bad}). It failed to use color effectively, placed incomparable variables on the same axis, and did not highlight the critical cost-effective outlier.

\begin{figure}
    \centering
    \includegraphics[width=1\linewidth]{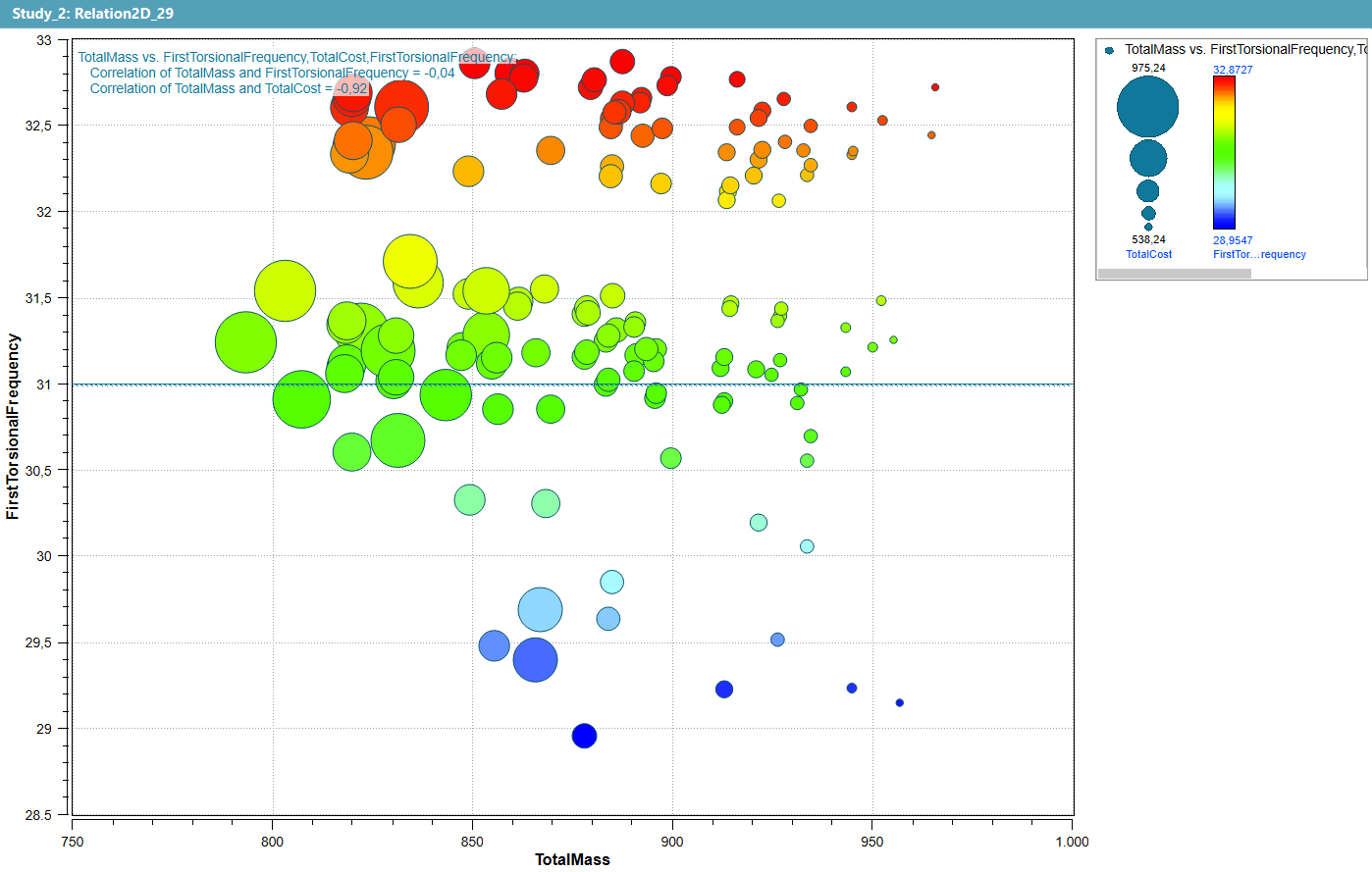}
    \caption{AI agent generated 2D relation plot}
    \label{fig:2d_relation}
\end{figure}

\textbf{Proposed System Result:} Agent generated an improved visualization (Fig. \ref{fig:2d_relation}) that highlights the optimal solution and enhances readability. While less insightful than expert plots in design set selection, it demonstrates clear improvements over the baseline through expert-derived visualization principles.

\subsubsection{Scenario: Creating an Informative Parallel Plot}

\textbf{Objective} A junior engineer requires a visualization to distinguish between the optimal design and outlier configurations in the dataset.

\textbf{User Input (provided to all systems)} \texttt{"Please generate a parallel plot."}

\begin{figure}[htbp]
    \centering
    \includegraphics[width=1\linewidth]{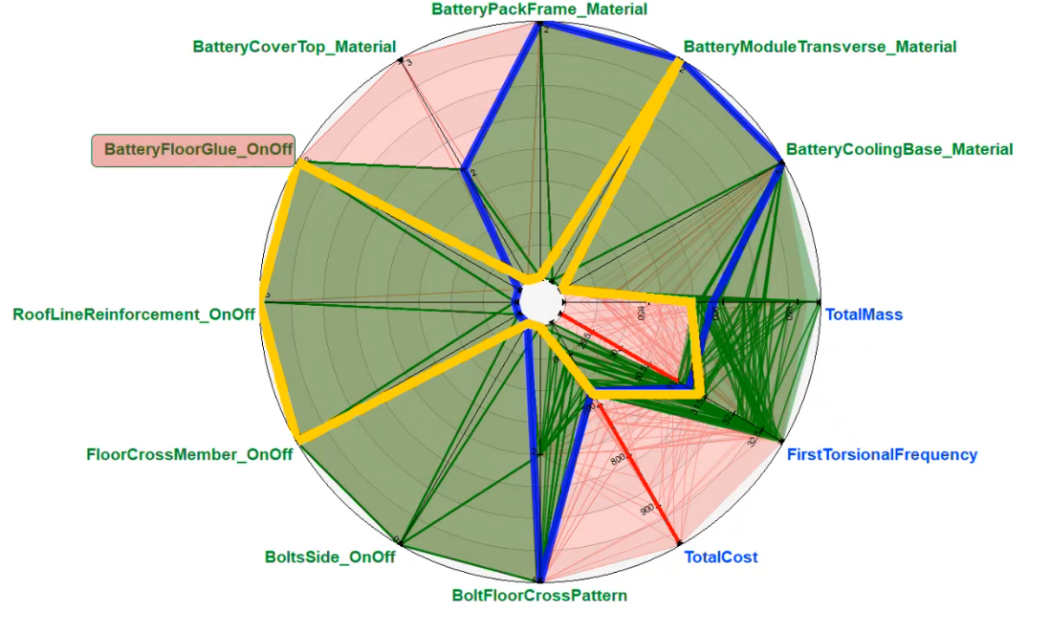}
    \caption{Human expert-generated parallel (radial) plot. Yellow: optimal design, Blue: outlier design, Green: feasible designs, Red: infeasible designs}
    \label{fig:humanradial}
\end{figure}

\textbf{Human Expert Result:} The expert visualization (Fig. \ref{fig:humanradial}) effectively reveals that the outlier uses aluminum for most components and lacks battery glue. While comprehensive, grouping range variables with categorical variables reduces readability.

\begin{figure}[htbp]
    \centering
    \includegraphics[width=1\linewidth]{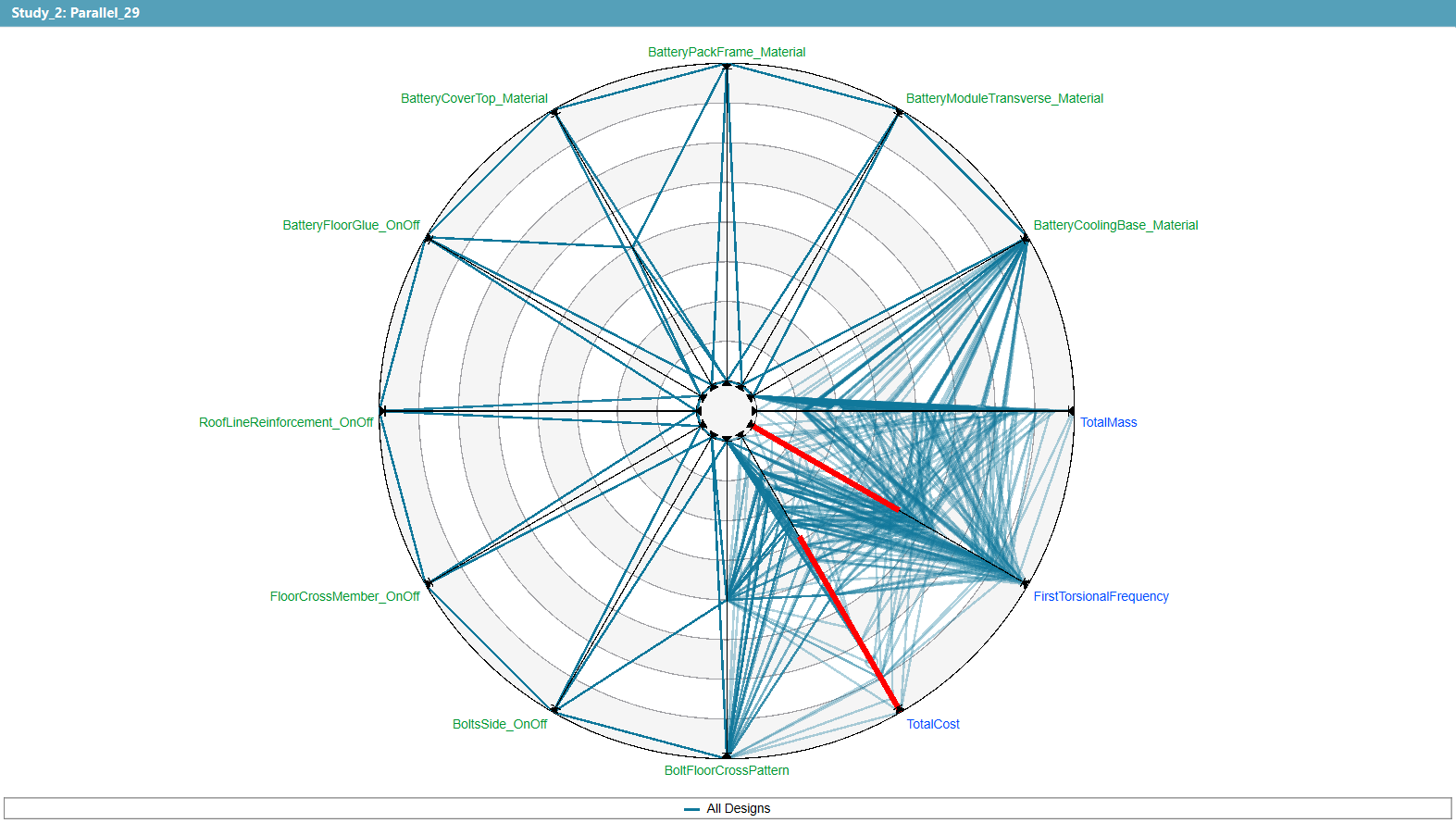}
    \caption{Baseline LLM+RAG generated parallel (radial) plot}
    \label{fig:baselineparallel}
\end{figure}

\textbf{Baseline System Result:} The baseline system produced functional code that indiscriminately includes all variables (Fig. \ref{fig:baselineparallel}), lacking emphasis on the optimal design and outlier, making it difficult to extract actionable insights.

\begin{figure}[htbp]
    \centering
    \includegraphics[width=1\linewidth]{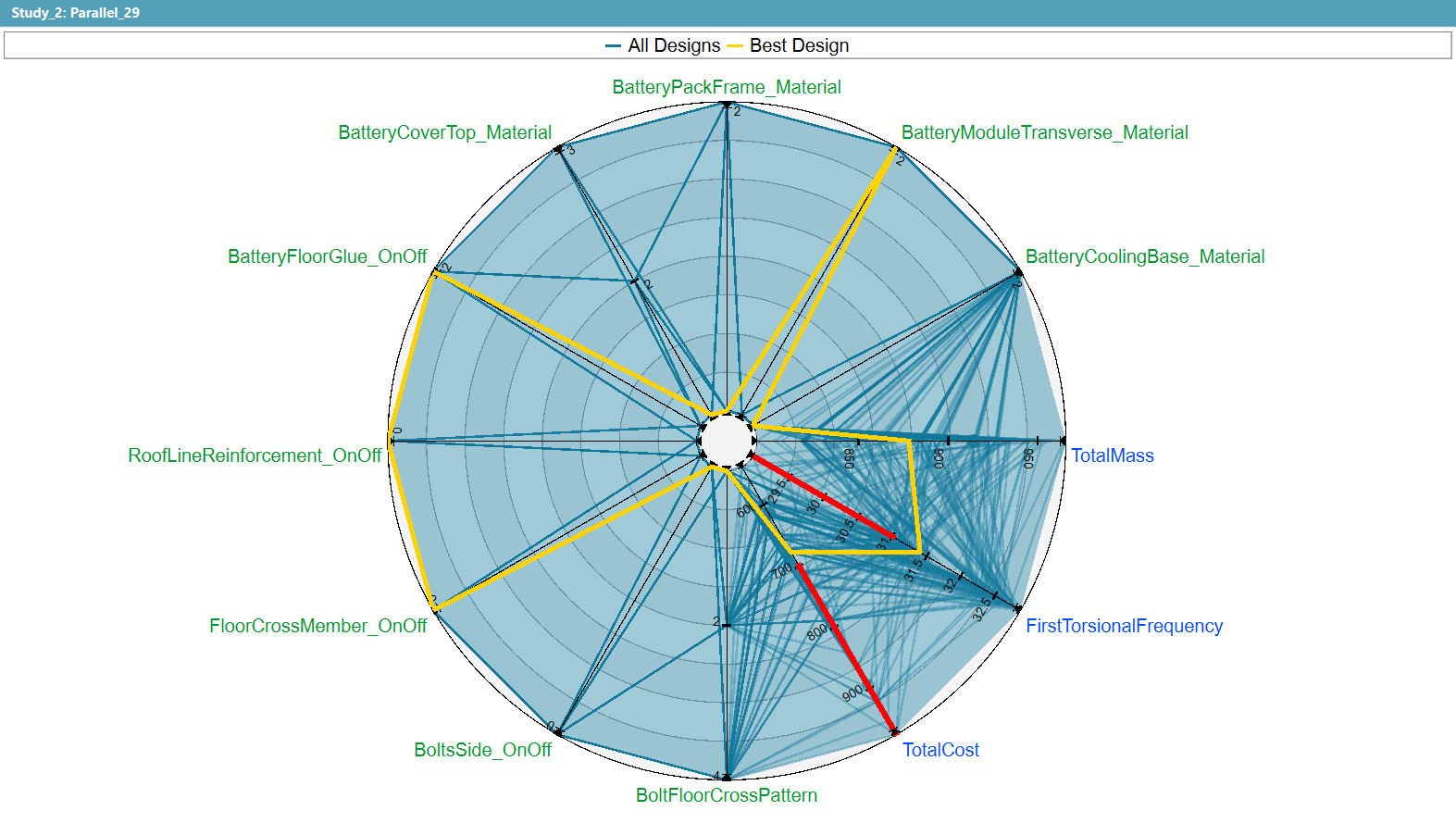}
    \caption{Proposed system-generated parallel (radial) plot with enhanced highlighting and legend}
    \label{fig:proposedparallel}
\end{figure}

\textbf{Proposed System Result:} Agent generated a more informative visualization (Fig. \ref{fig:proposedparallel}) with enhanced readability through optimal design highlighting, improved legends, and better visual hierarchy. However, it currently lacks outlier detection capabilities—a targeted area for future development.

\subsection{Quality of Generated Code}

The enhanced visualization quality directly correlates with the system's ability to generate superior code. We evaluated code quality using the methodology established by \cite{yeticstiren2023evaluating}, focusing on three critical dimensions: (1) Code validity - syntactic correctness and runtime error absence. (2) Code correctness - functional accuracy and adherence to requirements. (3) Output quality - effectiveness of resulting visualizations.

\textbf{Scope of Evaluation} We excluded code security from our assessment since the generated scripts serve solely visualization purposes and pose no security risks. Software reliability, as defined by \cite{kaur2014software}, was also omitted because our scripts execute once for visualization generation rather than requiring sustained operation. Similarly, code maintainability was not evaluated since these scripts are designed for single-use generation, though they can be adapted for reuse with appropriate modifications.

To address framework generalizability, we validated our physics-agnostic rules across three distinct engineering domains representing fundamentally different physical phenomena:

\subsubsection{Cross-Domain Validation: Project Descriptions}

Our validation employs five projects spanning three distinct engineering domains:

\textbf{Automotive Battery Design (Scripts 1-2):} Electrochemical optimization of battery systems for electric vehicles, focusing on parameters such as total mass, energy density, and thermal management. The optimization objectives include minimizing weight while maximizing capacity and first torsional frequency.

\textbf{Electric Motor Optimization (Script 4):} Electromagnetic motor design balancing competing objectives of torque output and efficiency while managing thermal constraints. Variables include rotor geometry, winding configurations, and material selections.

\textbf{Structural Control Arm Analysis (Script 5):} Mechanical optimization of automotive suspension control arms, focusing on structural stress analysis and weight reduction. Objectives include minimizing mass while maintaining structural integrity under load.

These three domains represent fundamentally different physical phenomena—electrochemical, electromagnetic, and mechanical systems—providing a rigorous test of the framework's physics-agnostic design.

\subsubsection{Evaluation Framework}

Initial testing revealed that standalone LLMs consistently generated Python-specific solutions using libraries such as matplotlib and seaborn, which are incompatible with the Simulation Analysis software environment. To ensure fair comparison, we evaluated our proposed agent approach against LLM+RAG (augmented with code examples).

\subsubsection{Expert Evaluation Process}

We conducted comprehensive testing across numerous examples, presenting five representative cases here. 12 evaluators with diverse technical backgrounds independently assessed the results: 
a simulation expert with 20 years of experience in CAE software;
a simulation optimization expert with multiple years of specialized domain knowledge; 
a software engineering manager with 12 years of experience in simulation software;
2 AI software engineering experts with extensive programming experience; 
a PhD in software engineering, leading a small data analytics and platform development team; 
a data scientist with 4 years of experience;
an electrical engineer with 2 years of experience in CAD design;
a computer scientist with 6 years of experience in coding;
a business economist with 1 year of experience in python;
a mechanical engineer with background in robotics and 4 years of experience in programming;
an electrical engineer with 1.5 years of experience in LLM development.

Additionally, the systems were evaluated by members of the product software team, but these results are proprietary and cannot be disclosed in this paper.

To enhance the robustness of our assessment methodology, we employed Claude 4.5 Sonnet as a third assessor to provide triangulation of findings and reduce potential human bias in evaluation. Additionally, we evaluated Claude's alignment with expert human judgment to assess whether the AI assessor accurately captures what expert humans consider to be superior solutions.

\subsubsection{Assessment Criteria}

\textbf{Code Validity} Generated scripts were evaluated for syntactic correctness using binary scoring (0 = invalid, 1 = valid). Following \cite{yeticstiren2023evaluating}, this measures compliance with programming language syntax rules and identifies potential runtime errors.

\textbf{Code Correctness} We assessed ``the extent to which the generated code performs as intended'' \cite{yeticstiren2023evaluating} across four dimensions using binary scoring based on \cite{almanasra2025analysis}:

\begin{itemize}
\item \textit{Code efficiency}: Optimization level without compromising functionality (0 = requires optimization, 1 = appropriately concise)
\item \textit{Documentation quality}: Adequacy of comments and documentation (0 = absent/inadequate, 1 = present/sufficient)
\item \textit{Exception handling}: Implementation of error handling mechanisms (0 = absent, 1 = present)
\item \textit{Code cleanliness}: Absence of unused variables or redundant elements (0 = contains unused elements, 1 = clean)
\end{itemize}

\textbf{Output Quality} We evaluated visualization effectiveness based on the script's ability to: (i) select appropriate data dimensions for analytical goals, (ii) use effective visual encoding for clear communication, and (iii) highlight critical information supporting decision-making.

\begin{table}[h]
\centering
\scriptsize
\setlength{\tabcolsep}{2.5pt}
\begin{tabular}{|l|c|c|c|c|c|c|c|}
\hline
\multirow{2}{*}{\textbf{Metric}} 
    & \multirow{2}{*}{\textbf{Scenario}}
    & \multicolumn{3}{c|}{\textbf{Human (n=12)}} 
    & \multicolumn{3}{c|}{\textbf{Claude 4.5 Sonnet}} \\
\cline{3-8}
    & & Mean & SD & Mode & Mean & SD & Mode \\
\hline
\multirow{5}{*}{{Code Validity}}
    & S1 & 1.00 & 0.00 & 1 & 1.00 & - & 1 \\
    & S2 & 1.00 & 0.00 & 1 & 1.00 & - & 1 \\
    & S3 & 1.00 & 0.00 & 1 & 1.00 & - & 1 \\
    & S4 & 1.00 & 0.00 & 1 & 1.00 & - & 1 \\
    & S5 & 1.00 & 0.00 & 1 & 1.00 & - & 1 \\
\hline
\multirow{5}{*}{{Code Correctness}}
    & S1 & 3.42 & 0.51 & 3 & 2.00 & - & 2 \\
    & S2 & 2.83 & 0.58 & 3 & 2.00 & - & 2 \\
    & S3 & 3.00 & 0.43 & 3 & 3.00 & - & 3 \\
    & S4 & 2.83 & 0.58 & 3 & 3.00 & - & 3 \\
    & S5 & 3.08 & 0.29 & 3 & 3.00 & - & 3 \\
\hline
\multirow{5}{*}{{Output Quality}}
    & S1 & 2.75 & 0.62 & 3 & 3.00 & - & 3 \\
    & S2 & 2.00 & 1.21 & 3 & 3.00 & - & 3 \\
    & S3 & 2.75 & 0.62 & 3 & 3.00 & - & 3 \\
    & S4 & 3.00 & 0.00 & 3 & 3.00 & - & 3 \\
    & S5 & 2.50 & 1.00 & 3 & 3.00 & - & 3 \\
\hline
\end{tabular}
\caption{Code Quality Assessment - AI Agent: Per-Scenario Breakdown (Human: 12 evaluators per scenario; Claude: single AI evaluation per scenario)}
\label{tab:pipeline_all_assessors}
\end{table}

\begin{table}[h]
\centering
\scriptsize
\setlength{\tabcolsep}{2.5pt}
\begin{tabular}{|l|c|c|c|c|c|c|c|}
\hline
\multirow{2}{*}{\textbf{Metric}} 
    & \multirow{2}{*}{\textbf{Scenario}}
    & \multicolumn{3}{c|}{\textbf{Human (n=12)}} 
    & \multicolumn{3}{c|}{\textbf{Claude 4.5 Sonnet}} \\
\cline{3-8}
    & & Mean & SD & Mode & Mean & SD & Mode \\
\hline
\multirow{5}{*}{{Code Validity}}
    & S1 & 1.00 & 0.00 & 1 & 1.00 & - & 1 \\
    & S2 & 1.00 & 0.00 & 1 & 1.00 & - & 1 \\
    & S3 & 1.00 & 0.00 & 1 & 1.00 & - & 1 \\
    & S4 & 1.00 & 0.00 & 1 & 1.00 & - & 1 \\
    & S5 & 1.00 & 0.00 & 1 & 1.00 & - & 1 \\
\hline
\multirow{5}{*}{{Code Correctness}}
    & S1 & 2.33 & 0.65 & 2 & 2.00 & - & 2 \\
    & S2 & 2.83 & 0.39 & 3 & 2.00 & - & 2 \\
    & S3 & 1.83 & 1.11 & 3 & 2.00 & - & 2 \\
    & S4 & 2.08 & 0.79 & 2 & 2.00 & - & 2 \\
    & S5 & 2.67 & 0.49 & 3 & 2.00 & - & 2 \\
\hline
\multirow{5}{*}{{Output Quality}}
    & S1 & 0.75 & 0.87 & 0 & 1.00 & - & 1 \\
    & S2 & 1.17 & 1.19 & 0 & 0.00 & - & 0 \\
    & S3 & 0.50 & 0.67 & 0 & 0.00 & - & 0 \\
    & S4 & 0.42 & 0.90 & 0 & 0.00 & - & 0 \\
    & S5 & 1.42 & 1.24 & 1 & 0.00 & - & 0 \\
\hline
\end{tabular}
\caption{Code Quality Assessment - LLM with RAG only: Per-Scenario Breakdown (Human: 12 evaluators per scenario; Claude: single AI evaluation per scenario)}
\label{tab:rag_all_assessors}
\end{table}

\subsubsection{Results}
Tables \ref{tab:pipeline_all_assessors} and \ref{tab:rag_all_assessors} summarize the evaluation scores for both approaches. Both systems demonstrated equivalent performance in code validity and correctness, confirming LLMs' general capability for producing syntactically correct scripts. However, the agent approach significantly outperformed the baseline in output quality and instruction adherence across all evaluation criteria.

\textbf{Script 1 (Parallel/Radial Plots)} Tables show Proposed (Mean = 2.75, SD = 0.62) vs. Baseline (Mean = 0.75, SD = 0.87) — a 267\% improvement. The Mode difference (3 vs. 0) indicates most evaluators found baseline output analytically worthless, while proposed system outputs were consistently rated highest quality. The proposed system distinguished optimal/baseline designs and filtered for feasible solutions. Claude's divergent assessment (Baseline = 1.00 vs. human = 0.82) illustrates AI tendency to credit functional execution despite poor interpretability.

\begin{figure}
    \centering
    \includegraphics[width=1\linewidth]{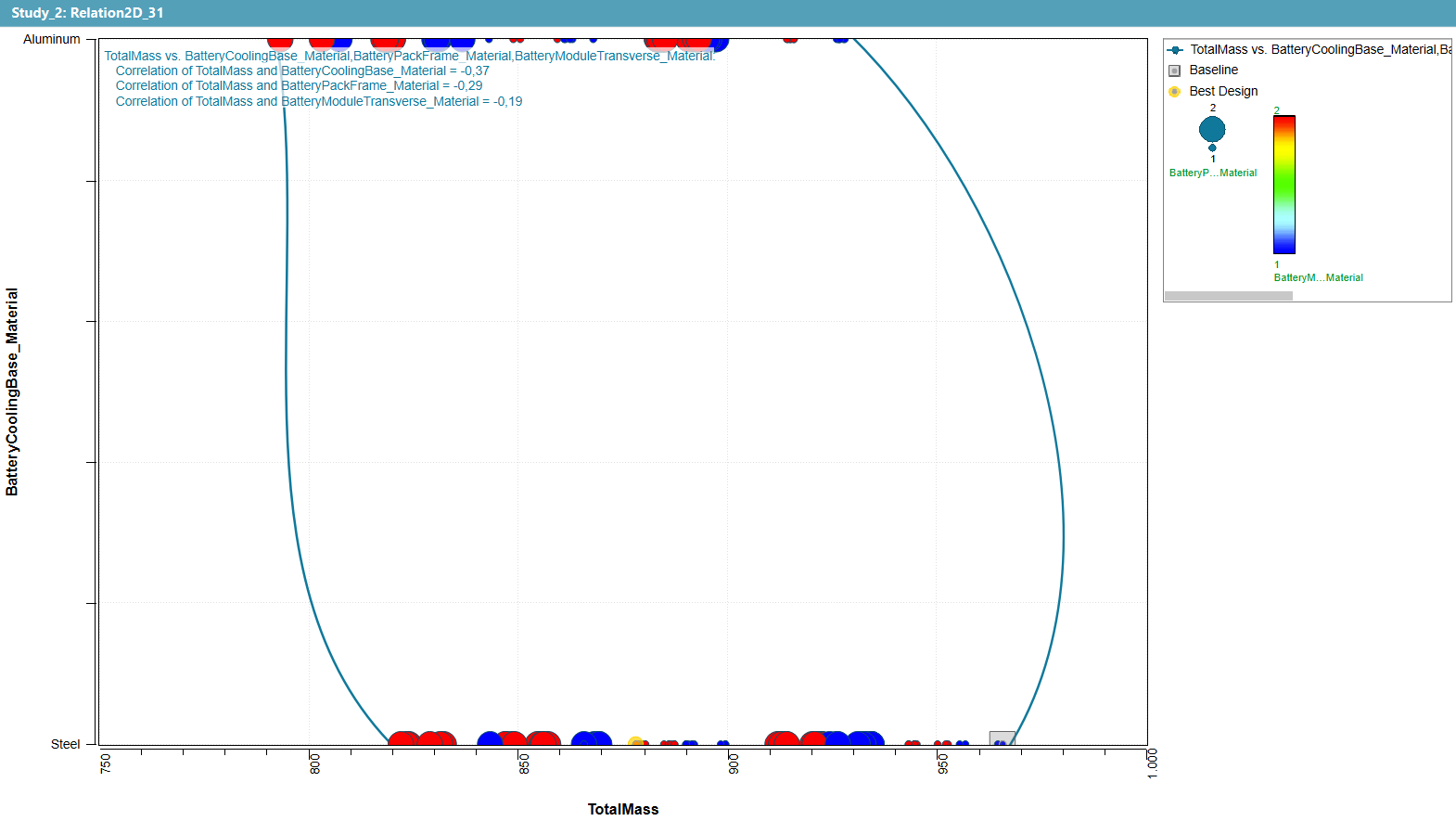}
    \caption{2D relation plot - Car Batteries - Proposed system}
    \label{fig:car_b}
\end{figure}

\textbf{Script 2 (2D Relation Plot)} Proposed system (Mean = 2.00, SD = 1.21, Mode = 3) vs. Baseline (Mean = 1.17, SD = 1.19, Mode = 0) demonstrates 71\% improvement. The baseline's Mode = 0 indicates most evaluators found minimal value, while the proposed system's Mode = 3 shows most evaluators rated it highest quality. The proposed system's variance (SD = 1.21) suggests some evaluators penalized suboptimal dimension selection (Fig. \ref{fig:car_b}) while others valued the correlation coefficients being displayed. Baseline's high variance (SD = 1.19) and Mode = 0 indicate fundamental disagreement on whether the output provided any value. Claude's assessment (Proposed = 3.00, Baseline = 0.00) aligned with human Mode patterns, correctly identifying the baseline's categorical-vs-continuous plotting error.

\begin{figure}
    \centering
    \includegraphics[width=1\linewidth]{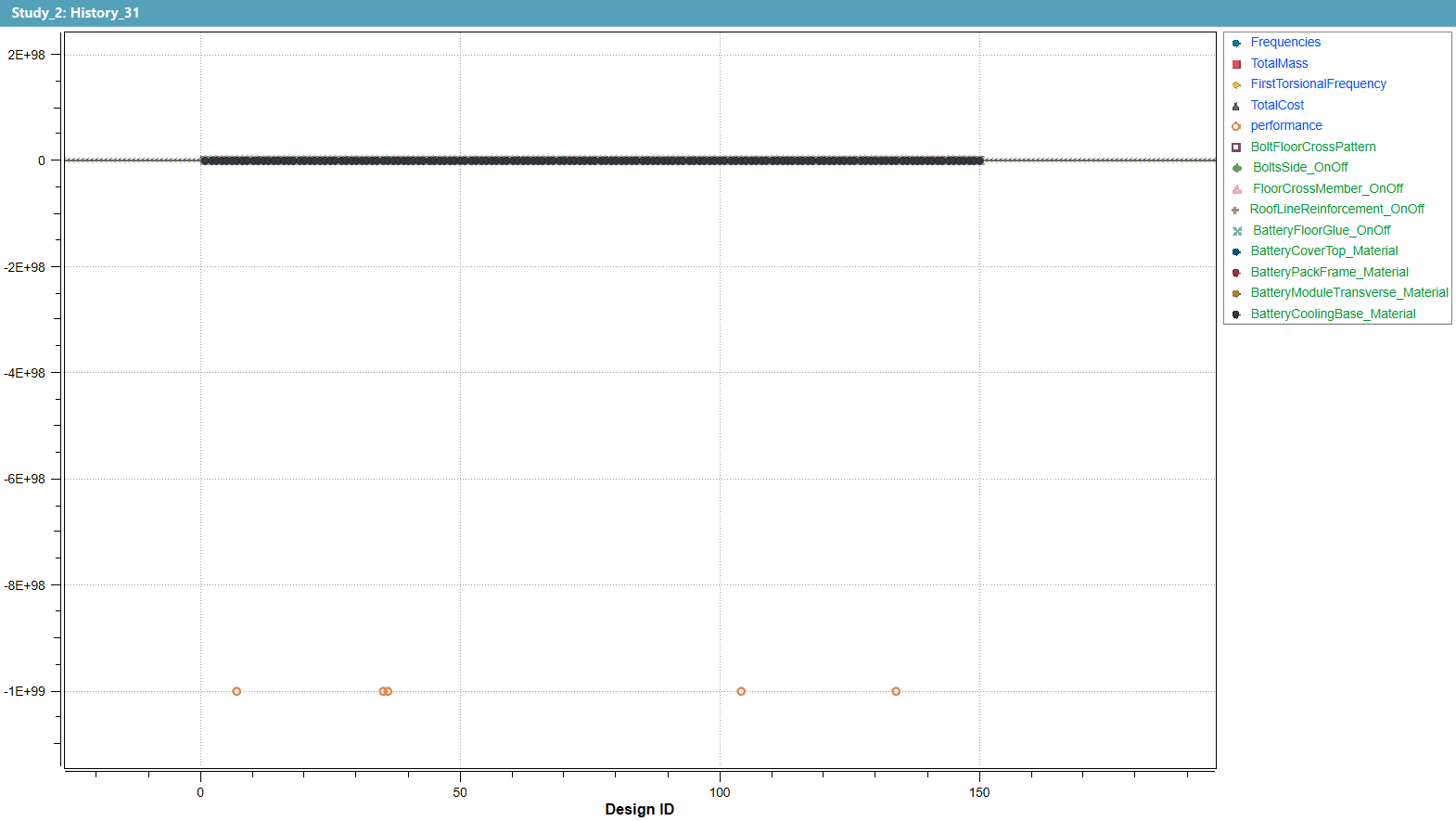}
    \caption{History plot - Car Batteries - Baseline system}
    \label{fig:his_bat_b}
\end{figure}

\textbf{Script 3 (History Plot)} Strong evaluator consensus for proposed system (Mean = 2.75, SD = 0.62, Mode = 3) versus baseline (Mean = 0.50, SD = 0.67, Mode = 0) demonstrates 450\% improvement. Baseline Mode = 0 (most evaluators rated it worthless) versus Proposed Mode = 3 (highest quality). The baseline (Fig. \ref{fig:his_bat_b}) plotted 14 parameters with catastrophic Y-axis scaling ($-1 E+99$ to $2E+98$), while proposed system applied normalization and convergence indicators (dashed lines)—directly using visualization expert rules from Section \ref{vizins}. Claude's perfect score (3.00) for proposed system aligns with human consensus, validating this specific rule implementation.

\textbf{Script 4 (History Plot - Electric Motors)} Achieved perfect human consensus for proposed system (Mean=3.00, SD = 0.00, Mode=3)—the only scenario with zero evaluator disagreement in output quality — versus baseline's poor performance (Mean = 0.42, SD = 0.90, Mode = 0). This demonstrates that when codified rules align perfectly with scenario requirements, the system achieves unanimous evaluators approval. The baseline repeated the critical flaw from Script 3, plotting 14 parameters simultaneously with catastrophic Y-axis scaling ($-1E+99$ to $2E+99$), creating an analytically worthless flat line. The proposed system intelligently selected related parameters (Torque and Efficiency), co-plotting them with dual Y-axes and convergence indicators. Claude's perfect score (3.00) matched human consensus, while rating baseline = 0.00 — more harshly than humans (1.00), suggesting AI assessors apply stricter standards for extreme scaling errors.

\begin{figure}
    \centering
    \includegraphics[width=1\linewidth]{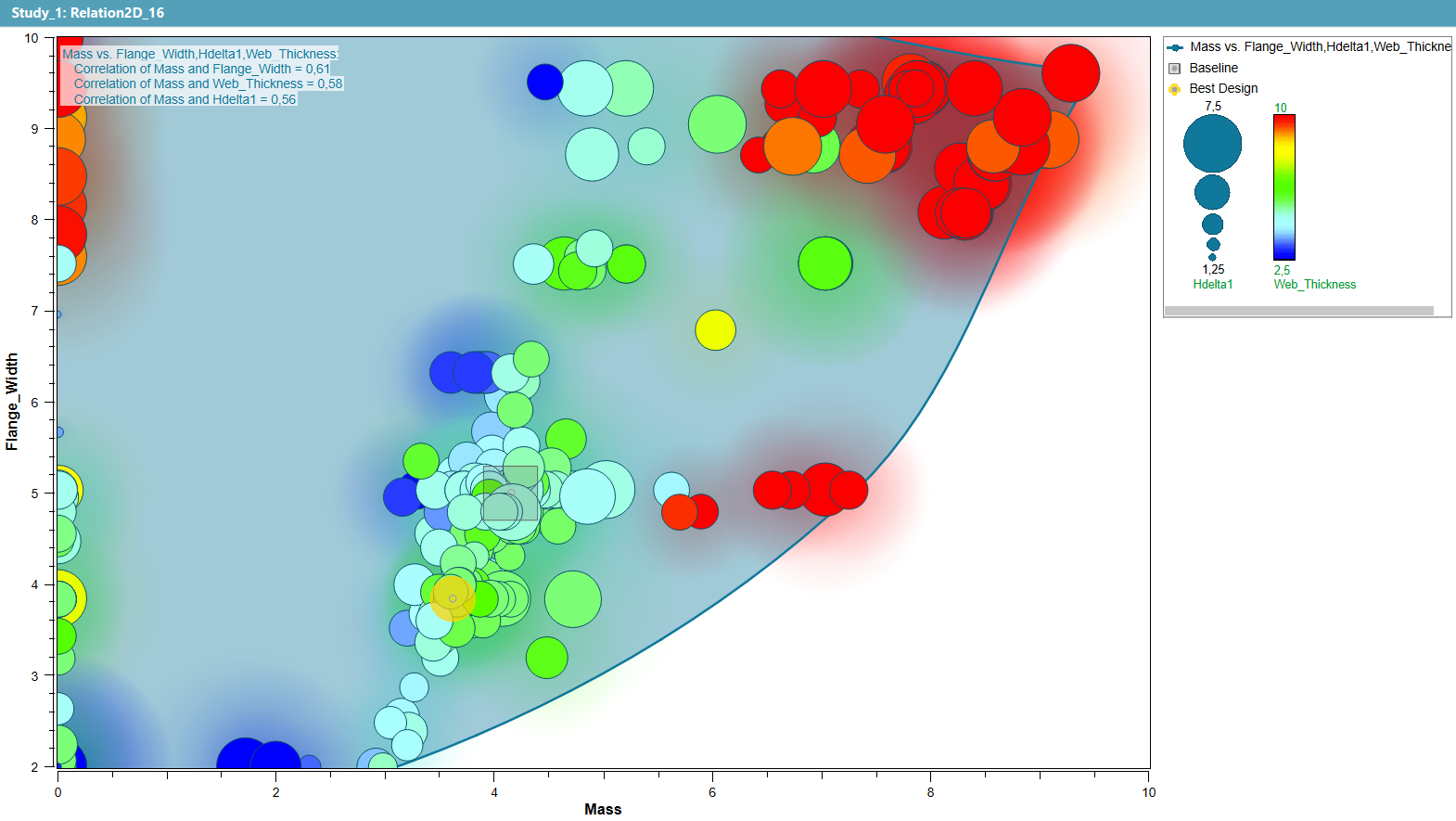}
    \caption{2D relation - Control Arms - proposed system}
    \label{fig:2drelationcontrol}
\end{figure}

\textbf{Script 5 (2D Relation Plot - Control Arms)} Baseline achieved its best Output Quality performance (Mean = 1.42, SD = 1.24, Mode = 1), yet still underperformed the proposed system (Mean = 2.50, SD = 1.00, Mode = 3), representing a 76\% improvement. The simulation expert praised the proposed system: "The plot clearly shows the optimization process, normalization along the Y-axis is done correctly, colors are chosen appropriately, and it is also convenient that the line type reflects convergence." The proposed system (Fig. \ref{fig:2drelationcontrol}) plotted Mass objective vs. Flange\_Width variable with multi-dimensional encoding (bubble size, color gradient, correlation coefficients: 0.61, 0.58, 0.55), while the baseline selected less informative variable-vs-response relationships. Claude rated the baseline = 0.00, more harshly than humans, noting it "creates a suboptimal visualization by plotting variable vs. response instead of the more analytically valuable objective vs. variable relationship, showing only a simple density cloud with limited optimization insights."

\subsection{Synthesis of Evaluation Results}

The technical evaluation demonstrates that our framework successfully codifies expert domain knowledge into an agent, addressing our three primary contributions: (1) systematic framework for capturing expert knowledge, (2) empirical validation, and (3) automated AI agent for addressing organizational bottlenecks.

\textbf{Contribution 1: Framework Validation}
The proposed system achieved 206\% higher Output Quality (mean: 2.60 vs. 0.85), with Mode = 3 across all scenarios versus baseline's Mode = 0 in 4/5 scenarios. This extends beyond \cite{vazquez2024llms}'s demonstration that LLMs can build charts, showing proper construction alone is insufficient without domain knowledge integration. In comparison to \cite{yang2024matplotagent}'s system of LLMs with prompts, our approach builds a single complex agent integrating a classifier, RAG, and codified rules. This unified architecture architecture enables potentially domain-agnostic rule application, validated across diverse physics simulation scenarios in this work.

Lower variance in Code Correctness (SD: 0.29-0.58 vs. 0.67-1.24) and Output Quality (SD: 0.00-1.00 vs. 0.67-1.24) demonstrates predictable results. Script 4's perfect consensus (SD = 0.00, Mean = 3.00) validates our unified framework achieves more predictable results than \cite{goswami2025plotgen}'s distributed multi-agent coordination.

\textbf{Contribution 2: Empirical Validation}
Expert assessments confirmed framework-codified principles translate into observable characteristics. The simulation expert noted baseline failures: "no proper scales and no normalization, rendering plots unreadable," while proposed outputs "clearly show the optimization process, normalization done correctly, colors chosen appropriately, and line type reflects convergence." While \cite{mallick2024chatvis} demonstrated LLM visualization using code snippets, our approach extends this by systematically codifying expert rules that address domain-specific analytical requirements beyond code examples.

Scripts 3 and 4 showed largest improvements (450\%, and 614\%) where baseline failed fundamental principles. Our 614\% improvement significantly exceeds \cite{zhu2023large}'s 30\% gain from rule integration, suggesting our comprehensive framework achieves synergistic effects beyond simple rule addition. Script 2's moderate improvement (71\%, SD=1.21) demonstrates our framework handles both primitive rules (normalization) and compositional rules (multi-dimensional encoding) effectively, aligning with \cite{wang2024can}'s observations about rule complexity.

AI assessor alignment (100\% on Code Validity) addresses \cite{mu2023can}'s earlier concerns about LLMs' rule-following—our results with newer versions demonstrate significantly improved adherence when rules are properly codified.

\textbf{Contribution 3: Addressing Expert Bottlenecks}
The framework's consistent Mode=3 ratings across diverse contexts demonstrate our AI agent successfully addresses challenges identified by \cite{grammel2010information} and \cite{Choe2024Enhancing} regarding non-experts' visualization struggles and expert interpretation requirements. Our approach goes beyond \cite{das2025charts}'s visualization literacy by autonomously generating expert-level outputs that non-experts create through simple prompts, enabling scalable production of high-quality visualizations without requiring domain expertise for each implementation.

\section{Conclusion}

This paper addresses the pervasive challenge of expert bottlenecks in organizations, where critical domain knowledge remains siloed with few specialists. We investigated how domain knowledge from human experts can be captured, codified, and leveraged to construct LLM-based AI agents capable of autonomous expert-level performance. We proposed and validated a systematic software engineering framework that empowers non-experts to achieve expert-level outcomes through AI agents.

We demonstrated our framework's efficacy through a rigorous case study in Simulation Analysis software simulation data visualization. We successfully engineered an AI agent by integrating a Retrieval-Augmented Generation (RAG) system for Simulation Analysis software-specific Python code generation, incorporating codified expert rules, and embedding visualization design principles directly into the Agent.

The key findings from our technical validation demonstrate the framework's effectiveness. Our agent achieved 206\% improvement in output quality (mean: 2.60 versus 0.85) across five scenarios spanning three simulation domains, with Mode = 3 ratings in all cases. 
Evaluator assessments validated the practical impact: baseline outputs were deemed unreadable with no proper scales and no normalization, while proposed system outputs received praise for showing the optimization process clearly with normalization done correctly and appropriate color choices. 
The system enables non-experts using simple prompts to generate visualizations that correctly apply nuanced expert rules—such as dashed lines for non-converged variables—that even human engineers sometimes miss. By codifying and embedding expert knowledge in this manner, our framework effectively bridges the expertise gap and mitigates the expert bottleneck, allowing non-experts to produce high-quality, insightful visualizations while freeing domain experts to focus on more complex, specialized tasks that truly require their unique skills. This addresses a critical organizational scalability challenge where essential domain knowledge traditionally remained concentrated among few specialists. 
To support this, a Physics-Agnostic design pattern proved essential for scalability to other domains. By decoupling visualization rules from specific physical phenomena, we achieved zero-shot reuse across battery, motor, and structural domains without retraining. 
Finally, strong AI-Human alignment suggests that 'LLM-as-a-Judge' frameworks can serve as reliable proxies for rapid regression testing to reduce human evaluation bottlenecks.

This research contributes a robust AI agent for visualization generation and a systematic, validated framework for engineering AI agents with human expert domain knowledge. Our solution represents a significant step towards democratizing access to specialized expertise through an agent, enabling more efficient and effective data analysis across industries, and ultimately fostering greater productivity and innovation within organizations. 

\section{Limitations and Threats to Validity}

\subsubsection{Limitations}

\textbf{Expert Selection and Internal Validity} Our framework development relied on knowledge from two experts within a single organization, potentially limiting the diversity of captured expertise and introducing selection bias. While methodologically appropriate for knowledge engineering \cite{hoffman1995eliciting}, this constrains the generalizability of codified rules.
\textbf{Evaluation Methodology} While code quality used standardized metrics 
\cite{yeticstiren2023evaluating}, visualization effectiveness required expert 
judgment with clear criteria, mitigated through multi-assessor triangulation.

\subsubsection{Scope and External Validity}

\textbf{Domain Specificity} Validation spanned three physics domains but remained 
within simulation-based optimization contexts. Effectiveness in non-simulation 
domains (medical visualization, financial analytics) remains unvalidated, limiting 
external validity.
\textbf{Organizational Context} The single-company case study within a large technology organization limits generalizability to different organizational contexts, user types, and expertise levels.

\subsection{Future Work}

Future research should address these limitations through: dynamic knowledge integration enabling learning from user feedback, enhanced user interaction capabilities for iterative refinement, cross-domain validation in multiple specialized fields, and large-scale evaluation through comprehensive user studies and standardized benchmarks.

\begin{acks}
The authors gratefully acknowledge the support provided by the Software Center. The authors also thank their colleagues and supervisors within Siemens for valuable discussions and feedback during this research. Special thanks to Roberto d'Ippolito, Ranny Sidhu, Jason Dunn and Guillermo Zschaeck from the product software team for their contributions, support and evaluation.
\end{acks}

\bibliographystyle{ACM-Reference-Format}
\bibliography{sample}
\end{document}